\documentclass{article}
\usepackage{graphicx} 
\usepackage[ruled]{algorithm2e}
\usepackage{amssymb}
\usepackage{amsmath}
\usepackage{mathtools}
\usepackage{todonotes}
\usepackage{url}

\newcommand{\Data}{\textbf{D}}
\newcommand{\FunctionFamily}{\textbf{F}}
\newcommand{\Coeffs}{\alpha}
\newcommand{\Labels}{\textbf{Y}}
\newcommand{\Evals}{\textbf{Z}}
\newcommand{\HyperParas}{\textbf{H}}
\newcommand{\KernelMat}{\textbf{K}_{\theta}}
\newcommand{\Parameters}{\Theta}
\newcommand{\Intermediate}{\tilde{g}}

\title{Algorithms for the Training of Neural Support Vector Machines}
\author{Lars Simon \\ Bundesdruckerei GmbH \\ {\tt lars.simon@bdr.de} \and Manuel Radons \\ Bundesdruckerei GmbH \\ {\tt manuel.radons@bdr.de}}
\date{August 2023}

\begin{document}

\maketitle

\begin{abstract}
    Neural support vector machines (NSVMs) allow for the incorporation of domain knowledge in the design of the model architecture.
In this article we introduce a set of training algorithms for NSVMs that leverage the Pegasos algorithm and provide a proof of concept by solving a set of standard machine learning tasks.
\end{abstract}

\section{Introduction}

The combination of neural networks (NNs) and support vector machines (SVMs) has been explored in numerous publications, see, e.g., \cite{Wiering2013NSVM}, \cite{Qi2016WhenEL}, or \cite{tang2015deep}.
A neural support vector machine (NSVM) is a machine learning model that combines (as the name would suggest) NNs and SVMs in its architecture.

A key benefit of this concept is that it allows to incorporate and exploit domain knowledge in the model architecture by adapting the NN part to a given data type. 
A straightforward sample application would be to use convolutional neural networks (CNNs) for the design of an NSVM image classifier.
The key drawback of NSVMs is that the SVM training runtime scales unfavourably with the number of samples in the training set. The Pegasos Algorithm addresses this problem \cite{Shalev2011Pegasos}. 
The basic idea of this article is to use the Pegasos algorithm to facilitate the NSVM training.

To this end we derive four NSVM training algorithms and test them on two standard data sets, Ringnorm and MNIST.
We would like to stress that at this point we do not claim any performance improvements over existing training procedures. The scope of this article is restricted to the derivation of the algorithms and a proof of concept. We feel, however, that our experimental results are encouraging and merit further investigation, both theoretical and pratical.

\subsection{Content and Structure}

In Sections \ref{sec:SVM} and \ref{sec:NSVM} we will introduce the Pegasos algorithm for SVMs and the concept of NSVMs, respectively. Section \ref{sec:algos} contains our main contribution, the derivation and description of four NSVM training algorithms that leverage the Pegasos algorithm.
Our proof of concept experiments are described in Section \ref{sec:experiments}, followed by some remarks on overfitting in Section \ref{sec:overfitting}. Section \ref{sec:conclusion} contains an outlook and our closing remarks.

\section{Support Vector Machines and Pegasos}
\label{sec:SVM}

	Let $n\in\mathbb{Z}_{\geq 1}$, $m\in\mathbb{Z}_{\geq 2}$. Assume we are given $(x_1 , y_1 ),\dots ,(x_m , y_m )\in \mathbb{R}^n\times\{-1,1\}$. This is interpreted as having $m$ datapoints $x_1 ,\dots , x_m\in\mathbb{R}^n$ divided into two classes, where $y_i\in\{-1,1\}$ determines the class of $x_i$, $i\in\{1,\dots ,m\}$. We furthermore assume that there exists at least one point in each of the two classes, i.e., $\{y_1 ,\dots , y_m\} = \{-1,1\}$. Given that we consider support vector machines without bias terms, we then consider the following optimization problem (here, $\lambda\in\mathbb{R}_{>0}$ is a hyperparameter):
	\begin{equation*}
		\begin{aligned}
			& {\text{minimize}}
			& & \frac{\lambda}{2}{\Vert w\Vert}^2 +  \frac{1}{m}\sum_{i=1}^{m}\max (0, 1 - y_i w^t x_i )\\
			& \text{subject to}
			& &  w\in \mathbb{R}^n .
		\end{aligned}
	\end{equation*}
	
	Let $K$ be a positive semidefinite kernel on $\mathbb{R}^n$, i.e., a map $K\colon\mathbb{R}^n\times\mathbb{R}^n\to\mathbb{R}$ with $K(a, b) = K(b, a)$ for all $a, b \in\mathbb{R}^n$, such that the matrix $\left(K(z_i , z_j )\right)_{1\leq i,j\leq l}$ is positive semidefinite, whenever $l\in\mathbb{Z}_{\geq 1}$ and $z_1 , \dots , z_l \in\mathbb{R}^n$. Let $(H, \langle\cdot ,\cdot\rangle_H)$ be the reproducing kernel Hilbert space associated to $K$. The corresponding kernelized version of the support vector machine minimization problem then becomes:
		\begin{equation*}
		\begin{aligned}
			& {\text{minimize}}
			& & \frac{\lambda}{2}{\Vert f\Vert}_H^2 +  \frac{1}{m}\sum_{i=1}^{m}\max \left(0, 1 - y_i f(x_i ) \right)\\
			& \text{subject to}
			& &  f\in H.
		\end{aligned}
	\end{equation*}
	
	While most approaches for training support vector machine revolve around solving the {\emph{dual}} problem, the Pegasos algorithm focusses exclusively on the primal. Because of this, we abstain from discussing the dual formulation of the support vector machine optimization problem.
	
	The Pegasos algorithm \cite{Shalev2011Pegasos} employs (stochastic) gradient descent on the objective of the primal problem and comes with various theoretical performance guarantees, e.g., with a favourable dependence of the runtime on the number of training samples. In the interest of brevity we only present the pseudocode for the kernelized version of the Pegasos algorithm (see Algorithm \ref{algo:originalpegasos}) and again refer to \cite{Shalev2011Pegasos} for a thorough derivation and discussion.

\setcounter{algocf}{-1}
\begin{algorithm}
		\SetKwInOut{Input}{Input}
		\SetKwInOut{Output}{Output}
		
		\Input{dataset $\{(x_1 ,y_1 ),\dots , (x_m , y_m )\}$, number of steps $T$, hyperparameter $\lambda >0$, positive semidefinite kernel $K$}
		\Output{coefficients $\alpha$}
		
		Initialization:
        Choose $i_1\in\{1,\dots , m\}$ uniformly at random.
        
        Let $\alpha^{(2)}\in\mathbb{R}^m$ with $\alpha_{i_1}^{(2)}=1$ and $\alpha_j^{(2)}=0$ all $j\in\{1,\dots ,m\}$ with $j\neq i_1$

        \For{t=2,\dots, T}
        {
Choose $i_t\in\{1,\dots , m\}$ uniformly at random

For all $j\in\{1,\dots ,m\}$ with $j\neq i_t$: set $\alpha_j^{(t+1)}=\alpha_j^{(t)}$

 \If{$\frac{y_{i_t}}{\lambda (t-1)}\sum_{j=1}^{m}\alpha_j^{(t)} y_j K(x_j , x_{i_t} ) < 1$}{
 $\alpha_{i_t}^{(t+1)} = \alpha_{i_t}^{(t)} + 1$

 }
        \Else{
       $\alpha_{i_t}^{(t+1)}   =\alpha_{i_t}^{(t)}$

        }

        }
       
  Set $\alpha\coloneqq\alpha^{(T+1)}$

		\Return $\alpha$
		
		\caption{Kernelized Pegasos Algorithm}
		\label{algo:originalpegasos}
	\end{algorithm}

\section{Neural Support Vector Machines}
\label{sec:NSVM}
Informally speaking, the Neural Support Vector Machine (NSVM) works by first computing relevant features of an input using a neural network and subsequently feeding the features computed by this procedure into a support vector machine (SVM), cf. \cite{Wiering2013NSVM}.
	
	More formally, let $\left(F_\theta \right)_{\theta\in\mathbb{R}^L}$ be a family of maps $F_\theta\colon\mathbb{R}^d\to\mathbb{R}^n$ and let $K$ be a positive semidefinite kernel on $\mathbb{R}^n$ with associated reproducing kernel Hilbert space $(H, \langle\cdot ,\cdot\rangle_H)$. For the sake of simplicity we further assume that the maps $F\colon\mathbb{R}^d\times\mathbb{R}^L\to\mathbb{R}^n$, $(x, \theta )\mapsto F_\theta (x)$ and $K$ are $\mathcal{C}^\infty$-smooth. In practice, $\left(F_\theta \right)_{\theta\in\mathbb{R}^L}$ is usually chosen to be a neural network, where $\theta$ represents the trainable parameters. The hypothesis class we consider is then the set of functions
	\begin{align*}
		\left\{f\circ F_\theta\colon \mathbb{R}^d\to\mathbb{R}\vert f\in H, \theta\in\mathbb{R}^L\right\}.
	\end{align*}
	Given an element $g$ in said hypothesis class, the associated classifier is given by the map $\mathbb{R}^d\to\{-1,1\}$ which maps $x\in\mathbb{R}^d$ to $1$ whenever $g(x)\geq 0$, and to $-1$ otherwise.
	
	Given data $(x_1 , y_1 ),\dots , (x_m , y_m )\in\mathbb{R}^d\times\{-1,1\}$ with $\{y_1 , \dots , y_m\} = \{-1, 1\}$, we choose a classifier from the aforementioned hypothesis class by solving an optimization problem. While the latter will differ throughout Algorithms 1-4 presented below, the basis for our considerations is the following minimization problem:
	\begin{equation}\label{nsvm_optimization_problem}
		\begin{aligned}
			& {\text{minimize}}
			& & \frac{\lambda}{2}{\Vert f\Vert}_H^2 +  \frac{1}{m}\sum_{i=1}^{m}\max \left(0, 1 - y_i f(F_\theta (x_i )) \right)\\
			& \text{subject to}
			& &  f\in H, \theta\in\mathbb{R}^L,
		\end{aligned}
	\end{equation}
	where, as before, $\lambda >0$ is a hyperparameter. The output of our algorithms will -- in all but one case -- be a map 
 \[
\mathbb{R}^d\to\mathbb{R}\,\quad
 x\mapsto \sum_{i=1}^m \alpha_i y_i K({F_\theta (x_i )}, {F_\theta (x)})
 \]	for some $\theta\in\mathbb{R}^L$ and $\alpha_1 ,\dots , \alpha_m \in \mathbb{R}_{\geq 0}$.

\section{Algorithms}
\label{sec:algos}
In this section we describe several algorithms for training a Neural Support Vector Machine. A thorough theoretical analysis of these algorithms is left for future work; in particular, we do not claim any performance guarantees. While the description of these algorithms may involve abstract elements of the reproducing kernel Hilbert space $H$, the respective implementations are elementary and make use of the kernel trick to forego calculations in $H$. 

	All algorithms take as an input a family of maps $\FunctionFamily\coloneqq \left(F_\theta \right)_{\theta\in\mathbb{R}^L}$, where $F_\theta\colon\mathbb{R}^d\to\mathbb{R}^n$, a positive semidefinite kernel $K$ on $\mathbb{R}^n$, a dataset \[\Data\coloneqq\{(x_1 , y_1 ),\dots , (x_m , y_m )\}\subset\mathbb{R}^d\times\{-1,1\}\] with $\{y_1 , \dots , y_m\} = \{-1, 1\}$, and the number of training steps $T\in\mathbb{Z}_{\geq 1}$. 
 Further, the inputs will include hyperparameters that vary between the algorithms. 
 For the sake of simplicity we assume that the maps $F\colon\mathbb{R}^d\times\mathbb{R}^L\to\mathbb{R}^n$, $(x, \theta )\mapsto F_\theta (x)$ and $K$ are $\mathcal{C}^\infty$-smooth. 

From the output of each algorithm we construct a function $g: \mathbb R^d\to\mathbb R$ that induces a classifier 
\[
  \mathbb{R}^d\to\{-1, 1\}\,,\quad
	x\mapsto
	\begin{cases}
		1 & \text{if }g(x)\geq 0\,,\\
		-1 & \text{if }g(x)<0\,.
	\end{cases}
\]
The construction method for $g$ varies from algorithm to algorithm.

We note that all algorithms will involve gradient descent steps wrt.\ the trainable parameters $\theta$. In practice, this may of course involve momentum, regularization etc., but, for the sake of brevity, we will not go into detail here.
Moreover, the vector of coefficients $\alpha$ will, in general, be sparse. For brevity we refrain from mentioning this fact in the pseudocode, even though its exploitation is a necessity in any sensible implementation.

\subsection{Algorithm 1}
\label{sec:algo_1}
In addition to the step number $T$, kernel $K$, dataset $\Data$ and function family $\FunctionFamily$ mentioned above, Algorithm 1 takes a hyperparameter $\lambda\in\mathbb{R}_{>0}$ as input.
Its output consists of vectors of corresponding coefficients $\alpha\coloneqq(\alpha_1 , \dots , \alpha_T )\in\left(\mathbb{R}_{\geq 0}\right)^T$, labels $\Labels\coloneqq\left(y_{i_1},\dots , y_{i_T}\right)\in\{-1,1\}^T$, and evaluations 
		$\Evals\coloneqq (z_1 , \dots , z_T )\in\mathbb{R}^{n\times T}$ -- as well as a vector of parameters $\Parameters\in\mathbb{R}^L$.
We use these to construct the aforementioned function $g$, from which our classifier derives, via  
 \[
 g(x)\ \coloneqq\ \frac{1}{\lambda T}\sum_{t = 1}^{T}\alpha_t y_{i_t} K(z_t , F_\Theta (x))\,.
 \]

A pseudocode for the training procedure is provided in Algorithm \ref{algo:1}. This algorithm is similar to the algorithm presented in \cite{gentinetta2023quantum}.

\subsubsection{Derivation of Algorithm}
 
	We now turn our attention to the optimization problem \ref{nsvm_optimization_problem}.
 Let $(H, \langle\cdot ,\cdot\rangle_H)$
 be the reproducing kernel Hilbert space associated to $K$
 and note that $H\times\mathbb{R}^L$ is a Hilbert space with the canonical inner product. Again, let $\Phi\colon\mathbb{R}^n\to H, x\mapsto K(x,\cdot )$ be the canonical feature map. We proceed in an analogous way to the original Pegasos paper. We initialize by setting $f_1 = 0\in H$ and randomly picking an initial value $\theta_1\in\mathbb{R}^L$. Then, in the $t$-th iteration, $t\in\{1,\dots ,T\}$, we assume that we are given $(f_t ,\theta_t )\in H\times\mathbb{R}^L$ from the previous iteration resp.\ from initialization. We then pick an index $i_t\in\{1,\dots ,m\}$ uniformly at random and consider the modified objective $C_t \colon H\times\mathbb{R}^L\to\mathbb{R}$ given by
	\begin{align*}
		(f, \theta )\mapsto & \mathrel{\phantom{=}}\frac{\lambda}{2}{\Vert f\Vert}_H^2 +  \max \left(0, 1 - y_{i_t} f(F_\theta (x_{i_t} )) \right)\\
		& = \frac{\lambda}{2}{\Vert f\Vert}_H^2 +  \max \left(0, 1 - y_{i_t}\langle f, \Phi (F_\theta (x_{i_t}))\rangle_H \right)
	\end{align*}
	If we have $y_{i_t}\langle f_t , \Phi (F_{\theta_t} (x_{i_t}))\rangle_H\neq 1$, then $C_t$ is (continuously) Fréchet differentiable in an open neighborhood of $(f_t ,\theta_t )$ and the partial Fréchet derivatives $\frac{\partial C_t}{\partial f}$ and $\frac{\partial C_t}{\partial \theta}$ exist and are continuous in an open neighborhood of $(f_t ,\theta_t )$. 
Note that we have to use Fréchet derivatives because, unlike in the original Pegasos paper, the span $\operatorname{span}_\mathbb{R}(\{\Phi (F_\theta (x_i ))\colon i\in\{1,\dots , m\}, \theta\in\mathbb{R}^L\})$, where $\Phi\colon\mathbb{R}^n\to H, x\mapsto K(x,\cdot )$ is the canonical feature map, might be an {\emph{infinite}} dimensional subspace of $H$.

 In this case, where $y_{i_t}\langle f_t , \Phi (F_{\theta_t} (x_{i_t}))\rangle_H\neq 1$, we have:
	\begin{align*}
		\frac{\partial C_t}{\partial f} (f_t ,\theta_t ) = \lambda\langle f_t ,\cdot \rangle_H - 
			\begin{cases}
				0 & \textbf{if }y_{i_t}\langle f_t , \Phi (F_{\theta_t} (x_{i_t}))\rangle_H > 1\\
				y_{i_t}\langle\Phi (F_{\theta_t} (x_{i_t})),\cdot\rangle_H & \textbf{if }y_{i_t}\langle f_t , \Phi (F_{\theta_t} (x_{i_t}))\rangle_H < 1
			\end{cases}.
	\end{align*}
	
	We now obtain $(f_{t+1}, \theta_{t+1})\in H\times\mathbb{R}^L$ from $(f_t ,\theta_t )$ via a gradient descent step. Note that gradient descent with respect to Fréchet derivatives makes sense in Hilbert spaces because of the Riesz representation theorem, cf. \cite{kantorovich2016functional}. Since $f$ and $\theta$ play different roles in the optimization problem and, in practice, we want to use momentum, regularization etc.\ for the $\theta$-update, we allow different step sizes for $f$ and $\theta$. Following the original Pegasos algorithm, we update $f_t$ with a step size of $1/(\lambda t)$, i.e.,
	\begin{align*}
		f_{t+1} & = f_t - \frac{1}{\lambda t}\frac{\partial C_t}{\partial f}(f_t , \theta_t )\\
		& = f_t - \frac{1}{\lambda t}\cdot
		\begin{cases}
			\lambda f_t & \textbf{if }y_{i_t}\langle f_t , \Phi (F_{\theta_t} (x_{i_t}))\rangle_H > 1\\
			\lambda f_t - y_{i_t}\Phi (F_{\theta_t} (x_{i_t})) & \textbf{if }y_{i_t}\langle f_t , \Phi (F_{\theta_t} (x_{i_t}))\rangle_H < 1
		\end{cases}\\
		& = \left(1-\frac{1}{t}\right) f_t + 
		\begin{cases}
			0 & \textbf{if }y_{i_t}\langle f_t , \Phi (F_{\theta_t} (x_{i_t}))\rangle_H > 1\\
			\frac{1}{\lambda t}\cdot y_{i_t}\Phi (F_{\theta_t} (x_{i_t})) & \textbf{if }y_{i_t}\langle f_t , \Phi (F_{\theta_t} (x_{i_t}))\rangle_H < 1
		\end{cases},
	\end{align*}
	where, by abuse of notation, we used the Riesz representation theorem to identify $\frac{\partial C_t}{\partial f}(f_t , \theta_t )$ (an element of the continuous dual of $H$) with the corresponding element of $H$.\\
	We now obtain $\theta_{t+1}\in\mathbb{R}^L$ from $\theta_t$ via a (gradient descent) step in the direction $-\frac{\partial C_t}{\partial \theta}(f_t ,\theta_t )$ (as mentioned above, we leave the step size and other details open). 

	Note that in the degenerate case, where $y_{i_t}\langle f_t , \Phi (F_{\theta_t} (x_{i_t}))\rangle_H = 1$, we update $(f_t , \theta_t )$ as in the case $y_{i_t}\langle f_t , \Phi (F_{\theta_t} (x_{i_t}))\rangle_H > 1$. In analogy to the original Pegasos paper, this approach can be justified by considering (Fréchet) subgradients.\\

\subsubsection{On the Kernel Trick}

	From the above derivation of Algorithm 1 it is not immediately clear how to implement it in practice without carrying out calculations in $H$, which might be computationally expensive or inaccessible.
We now describe how the algorithm can be carried out using only kernel evaluations, i.e., avoiding explicit calculations in $H$. 
 To this end, we introduce
	\begin{align*}
		\Intermediate_t & = \langle f_t , \Phi (F_{\theta_t} (x_{i_t}))\rangle_H ,\\
		\alpha_t & = \begin{cases}
			1 & \text{if }y_{i_t}\Intermediate_t < 1\\
			0 & \text{if }y_{i_t}\Intermediate_t \geq 1
		\end{cases}
	\end{align*}
	for all $t\in\{1,\dots , T\}$. An inductive argument now shows that, for all $t\in\{1,\dots ,T\}$, we have:
	\begin{align*}
	\Intermediate_t & = \begin{cases}
		0 & \text{if }t=1\\
		\frac{1}{\lambda (t-1)}\sum_{s=1}^{t-1}\alpha_s y_{i_s} K(F_{\theta_s}(x_{i_s}) , F_{\theta_t}(x_{i_t})) & \text{if }t\geq 2
	\end{cases}\in\mathbb{R},\\
	f_{t+1} & = \frac{1}{\lambda t}\sum_{s=1}^{t}\alpha_s y_{i_s} K(F_{\theta_s}(x_{i_s}) , \cdot )\in H.
	\end{align*}
	Furthermore, for all $t\in\{1,\dots , T\}$, the map $C_t (f_t ,\cdot )\colon\mathbb{R}^L\to\mathbb{R}$ is given by
	\begin{align*}
		\theta\mapsto & \mathrel{\phantom{=}}\frac{\lambda}{2}{\Vert f_t\Vert}_H^2 +  \max \left(0, 1 - y_{i_t}\langle f_t , \Phi (F_\theta (x_{i_t}))\rangle_H \right)\\
		& = \frac{\lambda}{2}{\Vert f_t\Vert}_H^2 + \begin{cases}
			0 & \textbf{if }y_{i_t}\langle f_t , \Phi (F_{\theta} (x_{i_t}))\rangle_H \geq 1\\
			1 - y_{i_t}\langle f_t , \Phi (F_\theta (x_{i_t}))\rangle_H & \textbf{if }y_{i_t}\langle f_t , \Phi (F_{\theta} (x_{i_t}))\rangle_H < 1
		\end{cases}\,.
	\end{align*}
	Hence, in the non-degenerate case we have 
 \[
\frac{\partial C_t}{\partial \theta}(f_t ,\theta_t )  = 0
 \]
 if $y_{i_t}\Intermediate_t > 1$ or $t=1$ and 
 \[
\frac{\partial C_t}{\partial \theta}(f_t ,\theta_t )  = \frac{\partial}{\partial \theta}\left({-\frac{y_{i_t}}{\lambda (t-1)}\sum_{s=1}^{t-1}\alpha_s y_{i_s} K( F_{\theta_s}(x_{i_s}) , F_\theta (x_{i_t}) )}\right) (\theta_t )
 \]
 if $y_{i_t}\Intermediate_t < 1$ and $t\geq 2$.
This allows us to compute $\frac{\partial C_t}{\partial \theta}(f_t ,\theta_t )$ in practice using automatic differentiation (recall the remark from above regarding the handling of the degenerate case). In case we do not have direct computational access to gradients of $K$, we can estimate them using a finite difference approximation (which is computationally expensive) or gradient estimators, for example the estimators featuring in the SPSA and RDSA algorithms. A similar comment was made in \cite{gentinetta2023quantum}. 
 
\subsubsection{Adjusting the Kernel}

We would like to avoid the situation where our $\theta$-update roughly corresponds to scaling $\Phi\circ F_\theta\colon\mathbb{R}^d\to H$ by a constant factor. While this might improve the value of the objective function of the optimization problem \ref{nsvm_optimization_problem}, it does not {\emph{qualitatively}} affect how well the classes can be separated. Because of this, under the additional assumption that $K(a,a)\neq 0$ for all $a\in\mathbb{R}^n$, one might optionally replace $K$ by the modified kernel $\tilde{K}\colon\mathbb{R}^n\times\mathbb{R}^n\to\mathbb{R}$ given by
	 \begin{align*}
	 	(a,b)\mapsto \frac{K(a,b)}{\sqrt{K(a,a)}\cdot\sqrt{K(b,b)}}
	 \end{align*}
	  -- which ensures that $\mathbb{R}^n$ gets mapped to the unit sphere in $H$ by $\Phi$.
   This approach is also applicable to Algorithm 2.
   \\

	\begin{algorithm}
		\SetKwInOut{Input}{Input}
		\SetKwInOut{Output}{Output}
		
		\Input{dataset \Data, number of steps $T$, hyperparameter $\lambda$, kernel $K$, family of functions $\FunctionFamily$}
		\Output{coefficients $\Coeffs$, labels $\Labels$, evaluations $\Evals$, parameters 
  $\Parameters$}
		
		Initialization: randomly pick $\theta_1\in\mathbb{R}^L$ and $i_1\in\{1,\dots , m\}$
  
  $\alpha_1 = 1$

  $z_1 = F_{\theta_1}(x_{i_1})$

  $\theta_2 = \theta_1$

        \For{t=2,\dots, T}
        {
Choose $i_t\in\{1,\dots , m\}$ uniformly at random
		
  $z_t = F_{\theta_t}
(x_{i_t})\in\mathbb{R}^n$

Compute \[\Intermediate_t = \frac{1}{\lambda (t-1)}\sum_{s=1}^{t-1}\alpha_s y_{i_s} K(z_s , z_t )\]

 \If{$y_{i_t}\Intermediate_t < 1$}{
 $\alpha_t = 1$

 Obtain $\theta_{t+1}\in\mathbb{R}^L$ from $\theta_t$ via a gradient descent step wrt.\ the minimization objective \[
 \mathbb{R}^L\to\mathbb{R},\quad \theta\mapsto -\frac{y_{i_t}}{\lambda (t-1)}\sum_{s=1}^{t-1}\alpha_s y_{i_s} K(z_s , F_\theta (x_{i_t}) )\]
 }
        \Else{
        $\alpha_t  =0$
        
		$\theta_{t+1} = \theta_t$
        }

        }
		
  Set $\alpha\coloneqq(\alpha_1 , \dots , \alpha_T )$

  Set $\Labels\coloneqq\left(y_{i_1},\dots , y_{i_T}\right)$

  Set $\Evals\coloneqq (z_1 , \dots , z_T )$

  Set $\Parameters\coloneqq\theta_{T+1}$

		\Return ($\alpha, \Labels, \Evals, \Parameters$)
		
		\caption{}
		\label{algo:1}
	\end{algorithm}

\subsection{Algorithm 2}
\label{sec:algo_2}
Algorithm 2 takes the same inputs $T$, $K$, $\Data$, $\FunctionFamily$ and $\lambda$ as Algorithm 1.  
The output consists of vectors $\alpha\in\left(\mathbb{R}_{\geq 0}\right)^m$, $\Evals\in\mathbb{R}^{n\times m}$, and $\Theta\in\mathbb{R}^L$ of coefficients, evaluations, and parameters, respectively.
From these we get $g:\mathbb R^d\to\mathbb R$ via
\[g(x)\ =\ \frac{1}{\lambda T}\sum_{j = 1}^{m}\alpha_j y_j K(z_j , F_\Theta (x))\,,\]
where the $y_i$ are the $m$ labels given in the data set $\Data$.

Algorithm \ref{algo:2} provides a pseudocode for the training procedure. 

\subsubsection{Derivation of Algorithm 2}

Key to the derivation of Algorithm 2 is the fact that, {\emph{for fixed $\tilde{\theta}\in\mathbb{R}^L$}}, an optimal point $f^*\in H$ for the optimization problem
		\begin{equation*}
		\begin{aligned}
			& {\text{minimize}}
			& & \frac{\lambda}{2}{\Vert f\Vert}_H^2 +  \frac{1}{m}\sum_{i=1}^{m}\max \left(0, 1 - y_i f(F_{\tilde{\theta}} (x_i )) \right)\\
			& \text{subject to}
			& &  f\in H,
		\end{aligned}
	\end{equation*}
	necessarily satisfies $f^*\in\operatorname{span}_\mathbb{R}(\{ K( F_{\tilde{\theta}} (x_i ) , \cdot ) \vert i\in\{1,\dots ,m\}\})$ due to the representer theorem.\\
 
	Since Algorithm 2 is quite similar to Algorithm 1, we only sketch the difference in the update procedure and refer to the pseudocode for a detailed description: 
 For step $t\in\{1,\dots T\}$, we first update $f_t$ with a gradient descent step as in Algorithm 1, to obtain some $\tilde{f}_{t+1}\in H$. Then $\theta_t$ is updated with a gradient descent step to obtain some $\theta_{t+1}\in\mathbb{R}^L$. Subsequently, we obtain $f_{t+1}$ by "approximately projecting" $\tilde{f}_{t+1}$ to $\operatorname{span}_\mathbb{R}(\{ K( F_{\theta_{t+1}} (x_i ) , \cdot ) \vert i\in\{1,\dots ,m\}\})$.
   Moreover, the objective for the update of $\theta_t$ was slightly adjusted (from the original objective in Algorithm 1) with the hope of ensuring better compatibility with the aforementioned approximate projection.

\begin{algorithm}
		\SetKwInOut{Input}{Input}
		\SetKwInOut{Output}{Output}
		
		\Input{dataset \Data, number of steps $T$, hyperparameter $\lambda$, kernel $K$, family of functions $\FunctionFamily$}
		\Output{coefficients $\Coeffs$, evaluations $\Evals$, parameters 
  $\Parameters$}
		
		Initialization: randomly pick $\theta_1\in\mathbb{R}^L$ and $i_1\in\{1,\dots , m\}$ 

  Let $\alpha^{(2)}\in\mathbb{R}^m$ with $\alpha_{i_1}^{(2)}=1$ and $\alpha_j^{(2)}=0$ all $j\in\{1,\dots ,m\}$ with $j\neq i_1$

  $\theta_2=\theta_1$

        \For{t=2,\dots, T}
        {
Choose $i_t\in\{1,\dots , m\}$ uniformly at random

For all $j\in\{1,\dots ,m\}$ with $j\neq i_t$: set $\alpha_j^{(t+1)}=\alpha_j^{(t)}$

Compute
		\[\Intermediate_t = \frac{1}{\lambda (t-1)}\sum_{j=1}^{m}\alpha_j^{(t)} y_j K(F_{\theta_t} (x_j ) , F_{\theta_t} (x_{i_t}) )
		\]

 \If{$y_{i_t}\Intermediate_t < 1$}{
 $\alpha_{i_t}^{(t+1)} = \alpha_{i_t}^{(t)} + 1$

 Obtain $\theta_{t+1}\in\mathbb{R}^L$ from $\theta_t$ via a gradient descent step wrt.\ the minimization objective
 \[\mathbb{R}^L\to\mathbb{R}\,,\quad \theta\mapsto -\frac{y_{i_t}}{\lambda (t-1)}\sum_{j=1}^{m}\alpha_j^{(t)} y_j K(F_{\theta} (x_j ) , F_{\theta} (x_{i_t}) )\]
 }
        \Else{
       $\alpha_{i_t}^{(t+1)}   =\alpha_{i_t}^{(t)}$
       
	$\theta_{t+1} = \theta_t$
        }

        }
        \For{j=1,\dots,m}{$z_j \coloneqq F_{\theta_{T+1}}(x_j )$}
  
  Set $\alpha\coloneqq\alpha^{(T+1)}$

  Set $\Evals\coloneqq (z_1 , \dots , z_m )$

  Set $\Parameters\coloneqq\theta_{T+1}$

		\Return ($\alpha, \Evals, \Parameters$)
		
		\caption{}
		\label{algo:2}
	\end{algorithm}

\subsection{Algorithm 3}
\label{sec:algo_3}
Algorithm 3 employs the technique of mini batching. Hence, in addition to $T$, $K$, $\Data$, $\FunctionFamily$ and $\lambda$, it takes a batch size $k\geq 2$ as input, which controls the size of the mini batches used during the training; furthermore, it takes an additional hyperparameter $\mu >0$ as input.  In order to avoid divisions by $0$ we also assume $K(a,a)\neq 0$ for all $a\in\mathbb{R}^n$.
As in the case of Algorithm 2, the output consists of vectors $\alpha\in\left(\mathbb{R}_{\geq 0}\right)^m$, $\Evals\in\mathbb{R}^{n\times m}$, and $\Theta\in\mathbb{R}^L$ of coefficients, evaluations, and parameters, respectively -- and also $g:\mathbb R^d\to\mathbb R$ is derived via the same formula
\[g(x)\ =\ \frac{1}{\lambda T}\sum_{j = 1}^{m}\alpha_j y_j K(z_j , F_\Theta (x)),\]
where the $y_i$ are the $m$ labels given in the data set $\Data$. 
The differences between Algorithms 2 and 3 lie in the derivation of the output, that is, the application of mini batching during the training and the more involved minimization objective that is utilized in the gradient descent step. 

Algorithm \ref{algo:3} provides a pseudocode for the training procedure. 

\subsubsection{Derivation of Algorithm 3}

	Assume that $K(a,a)\neq 0$ for all $a\in\mathbb{R}^n$. In principle, this algorithm is analogous to Algorithm 2. However, for each step $t\in\{1,\dots , T\}$ of the algorithm, we pick a batch of samples, rather than a single sample: We choose a subset $A_t$ of $\{1,\dots ,m\}$ with cardinality $k\in\mathbb{Z}_{\geq 2}$, $k\leq m$, uniformly at random. We update $f$ analogously to Algorithm 2 and to the batched version of the Pegasos algorithm from the original paper. Furthermore, when updating $\theta$ for $t\geq 2$, we instead consider the minimization objective $\mathbb{R}^L\to\mathbb{R}$ (again anticipating the impending projection step as in Algorithm 2): 
	\begin{align*}
		\theta\mapsto & \mu\cdot\frac{
			\sum_{i,j\in A_t} \alpha^{(t+1)}_i \alpha^{(t+1)}_j y_i y_j K(F_\theta (x_i ), F_\theta (x_j ))
		}
		{
			\sqrt{\sum_{i,j\in A_t} \left( \alpha^{(t+1)}_i \alpha^{(t+1)}_j\right)^2}
			\cdot
			\sqrt{\sum_{i,j\in A_t} K(F_\theta (x_i ), F_\theta (x_j ))^2}
		}\\
		& -
		\frac{
			\sum_{i,j\in A_t} y_i y_j K(F_\theta (x_i ), F_\theta (x_j ))
		}
		{
			k
			\cdot
			\sqrt{\sum_{i,j\in A_t} K(F_\theta (x_i ), F_\theta (x_j ))^2}
		}\,,
	\end{align*}
	where $\mu >0$ is a hyperparameter and $\alpha^{(t+1)}$ is as in Algorithm 2. If $\alpha^{(t+1)}_i = 0$ for all $i\in A_t$, then the first fraction is not well-defined. In this case we replace the first fraction by $0$. The first fraction represents a replacement for the $\Vert f\Vert^2_H$-term in the objective of the minimization problem \ref{nsvm_optimization_problem}, which is less computationally expensive to evaluate on the one hand and on the other hand encourages shrinking of the $\Vert f\Vert^2_H$-term by means other than merely shrinking the Frobenius norm of the matrix $\left({K(F_\theta (x_i ), F_\theta (x_j ))}\right)_{i,j\in A_t}$ (and, in particular, by means other than merely scaling $f$).
 	The second fraction in the objective for the $\theta$ update was introduced in order to encourage an impovement of the {\emph{kernel-target alignment}}, cf. \cite{Cristianini2001Alignment}.
  It can be viewed as a replacement for the hinge loss terms in the objective of the minimization problem \ref{nsvm_optimization_problem}.\\
	
 Note that -- at least in the case where $K$ is normalized, i.e., $K(a,a)=1$ for all $a\in\mathbb{R}^n$ -- adjusting $\theta$ so that the Frobenius norm of the kernel matrix $\left({K(F_\theta (x_i ), F_\theta (x_j ))}\right)_{1\leq i,j\leq m}$ shrinks can lead to severe overfitting. This is intuitively clear when considering the hypothetical worst-case, where the kernel matrix approaches the identity matrix.\\

\begin{algorithm}
		\SetKwInOut{Input}{Input}
		\SetKwInOut{Output}{Output}
		
		\Input{dataset \Data, number of steps $T$, hyperparameters $\lambda ,\mu$, kernel $K$, family of functions $\FunctionFamily$, batch size $k$}
		\Output{coefficients $\Coeffs$, evaluations $\Evals$, parameters 
  $\Parameters$}
		
		Initialization: randomly pick $\theta_1\in\mathbb{R}^L$
		
		Choose a subset $A_1$ of $\{1,\dots , m\}$ with cardinality $k$ uniformly at random
		
		Let $\alpha^{(2)}\in\mathbb{R}^m$ with $\alpha_{j}^{(2)}=\frac1k$ if $j\in A_1$ and $\alpha_j^{(2)}=0$ otherwise 
		
		$\theta_2=\theta_1$

        \For{t=2,\dots, T}
        {
Choose a subset $A_t$ of $\{1,\dots , m\}$ with cardinality $k$ uniformly at random

For all $j\in\{1,\dots ,m\}$ with $j\not\in A_t$: set $\alpha_j^{(t+1)}=\alpha_j^{(t)}$

For all $i\in A_t$ compute
		\[\Intermediate^{(i)}_t = \frac{1}{\lambda (t-1)}\sum_{j=1}^{m}\alpha_j^{(t)} y_j K(F_{\theta_t} (x_j ) , F_{\theta_t} (x_i ) ) 
		\]

\For{all $i\in A_t$}{
\If{$y_i \Intermediate^{(i)}_t < 1$}{$\alpha_{i}^{(t+1)} = \alpha_{i}^{(t)} + \frac{1}{k}$}
\Else{$\alpha_{i}^{(t+1)} = \alpha_{i}^{(t)}$}
}

Obtain $\theta_{t+1}\in\mathbb{R}^L$ from $\theta_t$ via a gradient descent step wrt.\ the minimization objective $\mathbb{R}^L\to\mathbb{R}$: 
					\begin{align*}
						\theta\mapsto & \mu\cdot\frac{
							\sum_{i,j\in A_t} \alpha^{(t+1)}_i \alpha^{(t+1)}_j y_i y_j K(F_\theta (x_i ), F_\theta (x_j ))
						}
					   {
					   	   \sqrt{\sum_{i,j\in A_t} \left( \alpha^{(t+1)}_i \alpha^{(t+1)}_j\right)^2}
					   	   \cdot
					   	   \sqrt{\sum_{i,j\in A_t} K(F_\theta (x_i ), F_\theta (x_j ))^2}
				   	   }\\
						& -
						\frac{
							\sum_{i,j\in A_t} y_i y_j K(F_\theta (x_i ), F_\theta (x_j ))
						}
						{
							k
							\cdot
							\sqrt{\sum_{i,j\in A_t} K(F_\theta (x_i ), F_\theta (x_j ))^2}
						}
					\end{align*}
					If $\alpha^{(t+1)}_i = 0$ for all $i\in A_t$, then the first fraction is not well-defined. In this case we replace the first fraction by $0$

}
        \For{j=1,\dots,m}{$z_j \coloneqq F_{\theta_{T+1}}(x_j )$}
  
  Set $\alpha\coloneqq\alpha^{(T+1)}$

  Set $\Evals\coloneqq (z_1 , \dots , z_m )$

  Set $\Parameters\coloneqq\theta_{T+1}$

		\Return ($\alpha, \Evals, \Parameters$)
		
		\caption{}
		\label{algo:3}
	\end{algorithm}

\subsection{Algorithm 4}
	This algorithm consists of two parts: First we train the neural network with the aim of improving the kernel alignment, then, subsequently, we fit a support vector machine. Since the parameters of the neural network are no longer being trained when we fit the support vector machine, we can choose {\emph{any}} algorithm for fitting a support vector machine, e.g. the Pegasos algorithm, or an algorithm making use of the dual formulation. In order to accomodate this freedom of choice, we will leave the second part of the algorithm somewhat open, which of course leads to a less strict description of the algorithm.

In addition to the familiar inputs $T$, $K$, $\Data$, $\FunctionFamily$, and $k$, Algorithm 4 takes a 
$\mathcal{C}^\infty$-smooth loss function $\mathcal{L}\colon [-1,1]\times [-1,1]\to \mathbb{R}$ for the kernel alignment and a set $\HyperParas$ of hyperparameters for the fitting of the support vector machine in the second part of the algorithm. As in Algorithm 3 we also have to assume that $K(a,a)\neq 0$ for all $a\in\mathbb{R}^n$ and $k\geq 2$.
The outputs consist of a set $\Theta$ of parameters for the neural net and a classifier \[h:\mathbb R^n\ \to\ \{-1,1\}\,.\]
The function $g: \mathbb R^d\to\mathbb R$ is then given by $g=h\circ F_\Theta$.

\subsubsection{Derivaton of Algorithm 4}

We again assume that $K(a,a)\neq 0$ for all $a\in\mathbb{R}^n$. 
	As in Algorithm 3, we use mini batches: In the first part of the algorithm, when training the neural network, for each step $t\in\{1,\dots ,T\}$, we choose a subset $A_t$ of $\{1,\dots ,m\}$ with cardinality $k\in\mathbb{Z}$, $2\leq k\leq m$, uniformly at random. We then update $\theta$ via a gradient descent step with respect to the objective $\mathbb{R}^L\to\mathbb{R}$ given by
	\begin{align*}
		\theta\mapsto -
		\frac{
			\sum_{i,j\in A_t} y_i y_j K(F_\theta (x_i ), F_\theta (x_j ))
		}
		{
			k
			\cdot
			\sqrt{\sum_{i,j\in A_t} K(F_\theta (x_i ), F_\theta (x_j ))^2}
		},
	\end{align*}
	i.e., we update $\theta$ with the aim of improving the {\emph{kernel-target alignment}}, cf. \cite{Cristianini2001Alignment}.  Owing to the Cauchy-Schwarz inequality, the objective takes values in $[-1,1]$; noting furthermore that we have equality in the Cauchy-Schwarz inequality if and only if the two vectors of consideration are linearly dependent, we could alternatively (and more generally) consider the objective $\mathbb{R}^L\to\mathbb{R}$ given by
	\begin{align*}
		\theta\mapsto \mathcal{L}\left(1,
		\frac{
			\sum_{i,j\in A_t} y_i y_j K(F_\theta (x_i ), F_\theta (x_j ))
		}
		{
			k
			\cdot
			\sqrt{\sum_{i,j\in A_t} K(F_\theta (x_i ), F_\theta (x_j ))^2}
		}\right),
	\end{align*}
	where $\mathcal{L}\colon [-1,1]\times [-1,1]\to \mathbb{R}$ is a $\mathcal{C}^\infty$-smooth loss function. Of course, not all such $\mathcal{L}$ will be suitable for our purposes, so one might want to impose additional assumptions on $\mathcal{L}$. For the sake of brevity, however, we refrain from further expanding upon this thought.\\
 
	Upon completion of the first part of the algorithm, we are left with some $\theta_{T+1}\in\mathbb{R}^L$. In the second part of the algorithm, we then merely fit a support vector machine using our algorithm of choice on the data \[(F_{\theta_{T+1}} (x_1 ), y_1 ), \dots ,(F_{\theta_{T+1}} (x_m ), y_m )\in\mathbb{R}^n\times\{-1,1\}\,.\]

\begin{algorithm}
		\SetKwInOut{Input}{Input}
		\SetKwInOut{Output}{Output}
		
		\Input{dataset \Data, number of steps $T$, kernel $K$, family of functions $\FunctionFamily$, batch size $k$, loss function $\mathcal{L}$, hyperparameters $\HyperParas$}
		\Output{classifier $h$, parameters 
  $\Parameters$}
		
		Initialization: Randomly pick $\theta_1\in\mathbb{R}^L$ 

\textbf{Part 1} (kernel alignment):

        \For{t=1,\dots, T}
        {
Choose a subset $A_t$ of $\{1,\dots , m\}$ with cardinality $k$ uniformly at random

Obtain $\theta_{t+1}\in\mathbb{R}^L$ from $\theta_t$ via a gradient descent step wrt.\ the minimization objective $\mathbb{R}^L\to\mathbb{R}$: 
				\begin{align*}
					\theta\mapsto \mathcal{L}\left(1,
					\frac{
						\sum_{i,j\in A_t} y_i y_j K(F_\theta (x_i ), F_\theta (x_j ))
					}
					{
						k
						\cdot
						\sqrt{\sum_{i,j\in A_t} K(F_\theta (x_i ), F_\theta (x_j ))^2}
					}\right)
				\end{align*}
}
\For{$j=1,\dots,m$}{Set $z_j := F_{\theta_{T+1}}(x_j )$}

\textbf{Part 2} (fit SVM):

Fit SVM with kernel $K$ on data $(z_1,y_1),\dots, (z_m,y_m)\in\mathbb R^n\times\{-1,1\}$, using $\HyperParas$ and obtain classifier  \[h\colon\mathbb{R}^n\to\{-1,1\}\]

  Set $\Parameters\coloneqq\theta_{T+1}$

		\Return ($h, \Parameters$)
		
		\caption{}
		\label{algo:4}
	\end{algorithm}

\section{Experiments}
\label{sec:experiments}
The purpose of our numerical experiments is to demonstrate the feasibility of the algorithms we described; we did not exhaustively benchmark the performance of our algorithms when compared to each other or to other well-known machine learning models. 
 
 A key benefit of NSVMs is that domain knowledge can be exploited by selecting the neural net architecture to best suit the problem at hand. 
 We showcase this approach in our  experiments. First, we train a fully connected neural network in combination with an RBF kernel on the Ringnorm dataset, cf. \cite{Ringnorm} and \cite{breiman2000ring}. Second, we train  a CNN in combination with an RBF kernel on the MNIST dataset, cf. \cite{lecun2010mnist}.

\subsection{Ringnorm Data Set}

The Ringnorm dataset consists of $7400$ samples with $20$ numerical features each, and poses a binary classification problem.
We randomly split the data into a training set with $6660$ samples and a testing set with $740$ samples; both training set and test set are approximately balanced with respect to the two classes.
$200$ samples from each class (so, a total of $400$ samples) were split off from the training set to serve as a validation set during hyperparameter tuning. After deciding on a hyperparameter configuration, we trained the models from scratch on the entire training set (i.e., including the $400$ samples that were previously split off).

We applied minimal preprocessing to the data. That is, we standardized each of the $20$ features by first subtracting its mean and subsequently dividing by its sample standard deviation (here, of course, mean and sample standard deviation were computed only with respect to the training set, and not with respect to the entire dataset). 

\subsubsection{Description of Models}

The kernel for all models was 
\[K\colon\mathbb{R}^{20}\times\mathbb{R}^{20}\to\mathbb{R}\,,\quad K(a,b) = \exp (-\Vert a-b\Vert^2 )\,.\]
The neural net consisted of a sequence of four fully connected layers with respective output dimensions $40, 30, 20, 20$; we used  a ReLU activation for each of these layers.
In case of Algorithms 1 and 2 the output was then normalized with respect to the Euclidean norm in $\mathbb{R}^{20}$ (where the denominator is artificially bounded from below by a small $\epsilon >0$ in order to avoid division by $0$ and to provide numerical stability); for further remarks on this topic, see Section \ref{sec:overfitting} that is concerned with the prevention of overfitting. 

\subsubsection{Training and Results}

All algorithms 
used gradient descent with momentum and $L^2$-regularization for the $\theta$-update (Algorithm 4 in Part I, the kernel alignment).
The hyperparameter $\lambda$ was set to $0.0001$ in all algorithms (in Algorithm 4 during Part II, the SVM fit via Pegasos). Algorithm 3 had an additional hyperparameter $\mu = 1$.
Batch sizes of Algorithms 3 and 4 were $16$. The loss function during Part I of Algorithm 4 was 
$\mathcal{L}(\beta , \gamma ) = (\beta - \gamma )^2$. 

Algorithms 1-4 achieved 98.1\% (80000 Steps), 97.6\% (70000 Steps), 97.0\% (4600 Steps), and 98.2\% (4200 Steps Part I, 33500 Steps Part II) accuracy, respectively. 
We note that a better performance could likely be achieved via hyperparameter tuning, but this would exceed the scope of the here presented proof of concept.

	\subsection{MNIST Data Set}

As we merely aimed for a proof of concept we restricted our experiments to the binary classification task of distinguishing the digits $0$ and $1$.
Adapting our algorithms to multiclass classification and regression problems is a straighforward task that we leave to future work.

Our training and testing set contain $12665$, resp., $2115$ $28\times 28$ grayscale images. Both are approximately balanced between digits $0$ and $1$.
$500$ images from each class (so, a total of $1000$ images) were split off from the training set to serve as a validation set during hyperparameter tuning. After deciding on a hyperparameter configuration, we trained the models from scratch on the entire training set (i.e., including the $1000$ images that were previously split off).

\subsubsection{Description of Models}

All models have two convolutional layers with $5\times 5$ filters, both of them followed by a $2\times 2$ max pooling layer and subsequent ReLU activation. Furthermore, we apply channel-wise dropout between the second convolutional layer and the subsequent max pooling layer. In each model the first convolutional layer has 10, the second one has 20 output channels.
For Algorithms 1 and 2 the output is then normalized
 with respect to the Euclidean norm in $\mathbb{R}^{320}$ (where the denominator is artificially bounded from below by a small $\epsilon >0$ in order to avoid division by $0$ and to provide numerical stability) and subsequently scaled by the factor $\sqrt{2}$; for further remarks on this issue, see Section \ref{sec:overfitting}.
 The kernel for Algorithms 1 and 2 is 
 \[K\colon\mathbb{R}^{320}\times\mathbb{R}^{320}\to\mathbb{R}\,,\quad K(a,b) = \exp (-\Vert a-b\Vert^2 )\,,\]
and for Algorithms 3 and 4 it is
\[K\colon\mathbb{R}^{320}\times\mathbb{R}^{320}\to\mathbb{R}\,,\quad K(a,b) = \exp (-\Vert a-b\Vert^2/320 )\,.\]

\subsubsection{Training and Results}

As before, all algorithms used gradient descent with momentum and $L^2$-regularization for the $\theta$-update (Algorithm 4 in Part I, the kernel alignment).
The hyperparameter $\lambda$ was set to $0.0001$ in all algorithms (in Algorithm 4 during Part II, the SVM fit via Pegasos). Algorithm 3 had an additional hyperparameter $\mu = 1$.
Batch sizes of Algorithms 3 and 4 were $4$ and $64$, respectively.
The loss function during Part I of Algorithm 4 was again
$\mathcal{L}(\beta , \gamma ) = (\beta - \gamma )^2$.

Algorithms 3 and 4 achieved 100\% classification accuracy (using 9500 steps and 400 steps Part I, 35000 steps Part II, respectively) on the test set; Algorithms 1 and 2 each misclassified exactly 2 samples from the test set (100000 steps each). We do not claim that this implies that our algorithms perform well, since this was an easy task. The point is that we can successfully use a CNN in our NSVM.

\section{Remarks on Overfitting}
\label{sec:overfitting}
Non-surprisingly, one needs to take measures to prevent overfitting in practice. While the neural support vector machine is a single machine learning model that should be considered as a whole, we aim to gain some insight into overfitting prevention by considering its two components (the neural network and the support vector machine) separately. With notation as in the previous sections, consider the kernel matrix
	\begin{align*}
\KernelMat\, \coloneqq\,		\left(K(F_\theta (x_i ), F_\theta (x_j ))\right)_{1\leq i,j\leq m}
	\end{align*}
	for a given $\theta$.\\
	
	One possible scenario is that the kernel matrix is well-behaved from the viewpoint of machine learning -- e.g., in the sense that the kernel-target alignment is large, or that the von Neumann entropy is neither close to $0$ nor close to $1$, c.f. \cite{Munoz2023Entropy} -- but the fixed choice of $\theta$ is still heavily overfit to the training data. That is, the kernel matrix might be well-behaved, while $F_\theta$ fails to compute the relevant features for previously unseen datapoints. In practice, this can be counteracted, e.g., with $L^1$- resp.\ $L^2$-regularization, or dropout.\\\
	
	Another scenario is that $\theta$ is not overfit to the training data, i.e. $F_\theta$ computes relevant features for previously unseen datapoints, but the kernel matrix $\KernelMat$ is not well-behaved.\\
	This second case is best illustrated with an example. Let $d=n$, and assume we are given a data set $(x_1 , y_1 ),\dots , (x_m , y_m )\in\mathbb{R}^d\times\{-1,1\}$
that contains both $-1$ and $1$ labeled entries. Further, let $F_\theta\colon\mathbb{R}^n\to\mathbb{R}^n$, $x\mapsto \theta x$ be the multiplication with a positive scalar $\theta\in\mathbb{R}_{>0}$. 
 Now set 
 \[
K\colon\mathbb{R}^n\times\mathbb{R}^n\to\mathbb{R}\,\quad K(z_1 ,z_2 )=\exp(-\Vert{z_1 -z_2}\Vert^2)\,.
 \]
 
 Assuming that $x_1 ,\dots ,x_m$ are pairwise distinct, we have 
	\begin{align*}
		\KernelMat = \left(\exp(-\theta^2\Vert{x_i -x_j}\Vert^2)\right)_{1\leq i,j\leq m} \xrightarrow{\theta\longrightarrow +\infty} I_{m\times m},
	\end{align*}
	and \begin{align*}
		\KernelMat = \left(\exp(-\theta^2\Vert{x_i -x_j}\Vert^2)\right)_{1\leq i,j\leq m} \xrightarrow{\theta\longrightarrow 0} \left(1\right)_{1\leq i,j\leq m},
	\end{align*}
	where $I_{m\times m}$ denotes the ${m\times m}$ identity matrix. Note that, of course, in this very simple example the case $\theta\xrightarrow{}\infty$ could be prevented using regularization, but that would be missing the point. Since $F_\theta$ is merely multiplication by a positive scalar, it does not {\emph{qualitatively}} change the dataset; in this sense $F_\theta$ does not fail to compute relevant features for previously unseen data. If $\theta$ is large, then the kernel matrix will be close to the identity matrix. The latter representes the overfitting regime: The training data is mapped to pairwise orthogonal vectors in the reproducing kernel Hilbert space associated to the kernel. If $\theta$ is small, then the kernel matrix will be close to the $m\times m$ matrix, for which every entry is equal to $1$. The latter represents the underfitting regime: All points from the set of training data are mapped to the same point in the reproducing kernel Hilbert space associated to the kernel.\\
	
	To prevent the second case from occurring, we need to ensure that the kernel matrix is well-behaved from the viewpoint of support vector machines. Noting that all four algorithms we presented are either already formulated with mini-batches or can easily be adapted to work with mini-batches, we could add a regularization term to the objective for the $\theta$-update, which penalizes the von Neumann entropy of the kernel matrix {\emph{for the current batch}} being close to either $0$ or $1$. In the case of non-negative normalized kernels, we could add a regularization term to the objective for the $\theta$-update, which penalizes the kernel matrix {\emph{for the current batch}} being close in Frobenius norm to either the identity matrix or the matrix for which all entries are equal to $1$. Furthermore, we can take measures that are specific to the kernel we are using. When using an RBF kernel, for example, one might choose to ensure that the feature vectors computed by $F_\theta$ are normalized, which will naturally bound the kernel values from below by a positive constant and thus ensure that the kernel matrix does not get too close to the identity matrix.\\
	Finally, we remark that the kernel alignment term featuring in Algorithms 3 and 4 offers some natural protection against the second case of overfitting we described.

\section{Conclusions and Outlook}
\label{sec:conclusion}

A key advantage of the NSVM concept is that it allows to incorporate domain knowledge in the design of the model architecture, e.g., by incorporating convolutional layers in the design of models for image classification.
The purpose of this article was to introduce a set of training algorithms for NSVMs that leverage the Pegasos algorithm and provide a proof of concept by solving a set of standard machine learning tasks. 
In particular, we demonstrated the feasibility of the training algorithms.

There are several possible alleys for future research.
Three of these stand out in our opinion: The move from binary classification to broader learning tasks such as multiclass classification and regression, a thorough theoretical analysis of the algorithms, e.g., with respect to convergence properties and computational costs, and the research of relevant use cases where our training methodology displays a tangible advantage over existing approaches.

Other directions are, building on the reflections of Section \ref{sec:overfitting}, the development of methods against overfitting, the incorporation of a bias term in the NSVM objective and experiments with other loss functions than the hinge loss.

\bibliographystyle{alpha}
\bibliography{references}
\end{document}